\documentclass[10pt,twocolumn,letterpaper]{article}
\usepackage[arxiv]{cvpr}   

\usepackage{soul}
\usepackage{xcolor}
\usepackage{multirow}
\usepackage{amssymb}%
\usepackage{pifont}%
\newcommand{\cmark}{\ding{51}}%
\newcommand{\xmark}{\ding{55}}%

\newcommand{\currentX}{\mathbf{X}_t} 
\newcommand{\currentXpixel}{X_t}
\newcommand{\referenceX}{\mathbf{X}_r} 
 
\newcommand{\alignedX}{\mathbf{X}_a} 

\newcommand{\motion}{\mathbf{M}} 
\newcommand{\vp}{\mathbf{p}}

\newcommand{\tp}[1]{{ {#1}}}

\definecolor{cvprblue}{rgb}{0.21,0.49,0.74}
\usepackage[pagebackref,breaklinks,colorlinks,citecolor=cvprblue]{hyperref}

\def\paperID{1710} 
\def\confName{CVPR}
\def\confYear{2024}

\DeclareMathOperator*{\argmin}{arg\,min}

\title{Enhancing Video Super-Resolution via Implicit Resampling-based Alignment}

\author{
Kai Xu\quad Ziwei Yu \quad  Xin Wang\quad Michael Bi Mi\quad Angela Yao
\\\vspace{1em} \texttt{kxu@comp.nus.edu.sg}
}

\begin{document}
\maketitle

\begin{abstract}
In video super-resolution, it is common to use a frame-wise alignment to support the propagation of information over time.  The role of alignment is well-studied for low-level enhancement in video, but existing works overlook a critical step -- resampling.  
We show through extensive experiments that for alignment to be effective, the resampling should preserve the reference frequency spectrum while minimizing spatial distortions. 
However, most existing works simply use a default choice of bilinear interpolation for resampling even though bilinear interpolation has a smoothing effect and hinders super-resolution.
From these observations, we propose an implicit resampling-based alignment.  
The sampling positions are encoded by a sinusoidal positional encoding, while the value is estimated with a {coordinate network} and a window-based cross-attention. 
\tp{We show that bilinear interpolation inherently attenuates high-frequency information while an MLP-based coordinate network can approximate more frequencies.}
Experiments on synthetic and real-world datasets show that alignment with our proposed implicit resampling enhances the performance of state-of-the-art frameworks with minimal impact on both compute and parameters.
\end{abstract}

\vspace{-1em}
\section{Introduction}\label{sec:intro}

Video super-resolution (VSR) recovers a high spatial resolution sequence of frames from a low-resolution sequence.  While image super-resolution can be applied naively to each frame individually, the temporal correlations across the frames give an extra source of information to improve the super-resolved output.  As such, the main difference in video versus image super-resolution architectures lies in the use of temporal dependencies. Previous works \cite{shi2022rethinking, chan2021basicvsr, wang2019edvr, haris2019recurrent} have shown that spatial alignment is an essential pre-processing step for effective information exchange across the frames.  Given the frame-to-frame camera and object motions, alignment provides indications of sub-pixel information which can benefit the super-resolution.

Frame-wise alignment estimates and compensates for motion. 
Motion estimation determines pixel displacements based on optical flow or additional offset prediction networks~\cite{liang2022recurrent_rvrt, chan2022basicvsrpp, wang2019edvr}. Motion compensation warps the reference to be aligned with the current frame.  During compensation, resampling is necessary because the warping may require non-discrete pixel values which are not present in the reference image.

\begin{figure}
\includegraphics[width=\linewidth]
{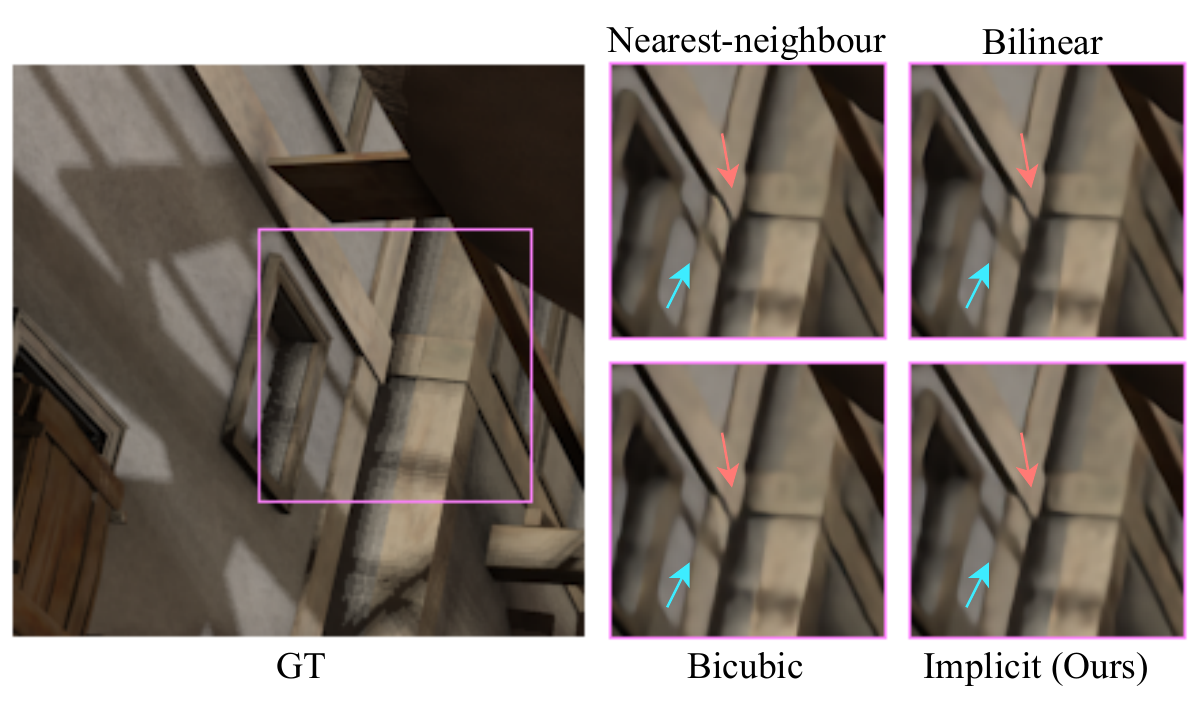}

\caption{Comparisons with super-resolved outcomes employing nearest-neighbor interpolation, bilinear and bicubic resampling. The red arrow highlights smoothing effects for bilinear and bicubic interpolation, while the blue arrow highlights the ragged edge.}\label{fig:preliminary_illustrations}
\vspace{-1em}
\end{figure}

Alignment is well-studied in low-level vision~\cite{lin2022unsupervised_S2SVR, chan2021basicvsr, xue2019video_tof}, but the role of resampling in alignment has been overlooked.  In fact, almost all existing works \cite{chan2021basicvsr,chan2022basicvsrpp,wang2019edvr,liang2022recurrent_rvrt} use a default bilinear interpolation due to its simplicity. 
Yet resampling is a critical step of alignment which should not be overlooked. As~\cref{fig:preliminary_illustrations} shows, the choice in resampling method can greatly impact the output.  
Resampling with bilinear and bicubic interpolation preserves the spatial structures of the original image, but tends to smooth out the intensity values.  Resampling with nearest-neighbour interpolation gives sharper results, albeit with spatial distortions and ragged edges.

To our knowledge, we are the first to investigate resampling in alignment for super-resolution; we take a deep dive and show the significant impact it can have.
The distinctions between resampling methods, particularly their impact on frequency reconstruction for sub-pixel values, \tp{become more evident when estimated motion can provide accurate sub-pixel offsets, that is, when the flow algorithms are more precise. As the resampling accuracy is hard to evaluate separately from the motion estimation accuracy, we examine the performance of resampling methods under ideal optical flow conditions using a synthetic dataset. This is the first study to isolate the effect of the resampling strategy with fixed flow in both synthetic and real-world settings.
} 
Our findings show that for a resampling method to be effective in alignment, it should avoid quantization in the coordinate transform and refrain from imposing low-pass filtering on the original signal.

Inspired by recent image implicit representations \cite{chen2021learning_liif, xu2021ultrasr}, we {propose a new alignment module} with an \emph{implicit} resampling. The resampling is achieved through a coordinate network with an local cross-attention module, applied to a feature window based on the motion offset.  Rather than explicitly interpolating on the reference frame for the sub-pixel feature value, we aggregate reference values with an affinity matrix based on the feature and positional encoding similarity.  Such an aggregation does not impose any smoothness constraints on the resampling process.  It also avoids spatial distortions by encoding the sub-pixel coordinate information into the sinusoidal positional encoding.  Consequently, our implicit resampling-based alignment module significantly outperforms both the state-of-the-art bilinear resampling-based alignments~\cite{chan2022basicvsrpp,liang2022vrt_vrt,liang2022recurrent_rvrt} and the nearest-neighbour resampling-based alignment~\cite{shi2022rethinking}.

Our proposed implicit resampling-based alignment once learned, can be applied across diverse alignment scenarios. In comparison, alignment modules in competing methods using deformable convolution~\cite{wang2019edvr} and deformable attention~\cite{liang2022recurrent_rvrt} must be learned specifically for fixed feature scales and alignment configurations. Our implicit resampling-based alignment is trained to handle \emph{all} feature scales and alignment configurations, enhancing generalization and reducing parameter size. 
We summarize our contribution as follows:
\begin{itemize}
    \item We highlight the previously overlooked role of resampling in alignment. Our studies show that effective resampling methods should both preserve the frequency spectrum while limiting spatial distortions.
    \item We propose an implicit resampling-based alignment method, where features and estimated motion are jointly learned through coordinate networks, and alignment is performed implicitly through window-based attention. Our implicit resampling-based alignment, once trained, can generalize to any feature scales and alignment configurations.
    \item Our proposed implicit resampling-based alignment surpasses current state-of-the-art alignment methods on video super-resolution tasks for both synthetic and real-world datasets, using either CNNs or Transformers as the backbone models.
\end{itemize}

\section{Related Work}
\label{sec:related_work}

\noindent\textbf{Image Resampling}
Aligning the reference frame to the destination frame requires resampling sub-pixel values on the discrete reference image. 
 Nearest-neighbour resampling directly looks up the values of nearest-neighbours; it is simple, but also has choppy distortions.
Smoother results can be achieved with bilinear (or bicubic) interpolation, which guarantees an L0- (or higher-order) smoothness on the resulting image intensity surface~\cite{dodgson1992image}. While smooth, the results are not edge preserving and as a result, can also be blurry~\cite{wolberg1990digital}. Recently, implicit representations in the form of neural networks have been proposed for encoding scenes~\cite{mildenhall2021nerf} and images~\cite{chen2021learning_liif}. 
~\cite{li2021neural_scene_flow_prior} leverage coordinate network as a prior for scene flow regularization. Our method shares the same insight, where we model the entire resampling and alignment process with coordinate networks and cross-attention mechanism.

\noindent\textbf{Video Super-Resolution}
Video super-resolution recovers a spatially high-resolution sequence from low-resolution frames. Its difference with image super-resolution lies in the use of temporal information. Early methods \cite{huang2017video, fuoli2019efficient, isobe2020video, isobe2020revisiting} did not consider spatial alignment from frame to frame.  Initially, VSR methods applied optical flow-based warping to 
align the neighbouring image inputs~\cite{kim2018spatio, xue2019video_tof}.  However, inaccurate flows lead to degradation 
and more recently, strategies have shifted either to align feature maps instead of images~\cite{chan2021basicvsr} or use the flow to guide deformable convolutions~\cite{wang2019edvr, chan2022basicvsrpp, liang2022vrt_vrt} and deformable attention schemes~\cite{liang2022recurrent_rvrt}.  To increase the robustness toward inaccurate optical flow, ~\cite{shi2022rethinking} propose patch alignment; they align blocks by averaging the motions within predefined grids.  We also consider a patch (referred to as a window in our work) context for cross-attention. However, our strategy differs as our window is dynamic, \ie each pixel's reference window is determined by its optical flow.

\noindent\textbf{Spatial \& Temporal Super-resolution} Image super-resolution aims to provide an up-sampled image from the low-resolution image and serves as the basis for video super-resolution. The recent work \cite{chen2021learning_liif, xu2021ultrasr} proposed to learn a continuous representation from the discrete image with an MLP.  
Video frame interpolation can be seen as a form of temporal super-resolution.  The interpolated frame is aggregated from adjacent frames by alignment and propagation. \cite{xue2019video_tof} 
Recently, ~\cite{DBLP:conf/cvpr/Niklaus020/Softmax_Splatting} proposed softmax splatting based on softmax resampling for interpolating frames in time. In this work, the resampling weights are related to the depth mask and the resampled value is based on relative occlusions. In contrast, our framework encodes sub-pixel information into positional encodings, reconstructing content at a higher frequency for VSR.

 \begin{figure*}[t]
\centering
\includegraphics[width=1.04\textwidth]{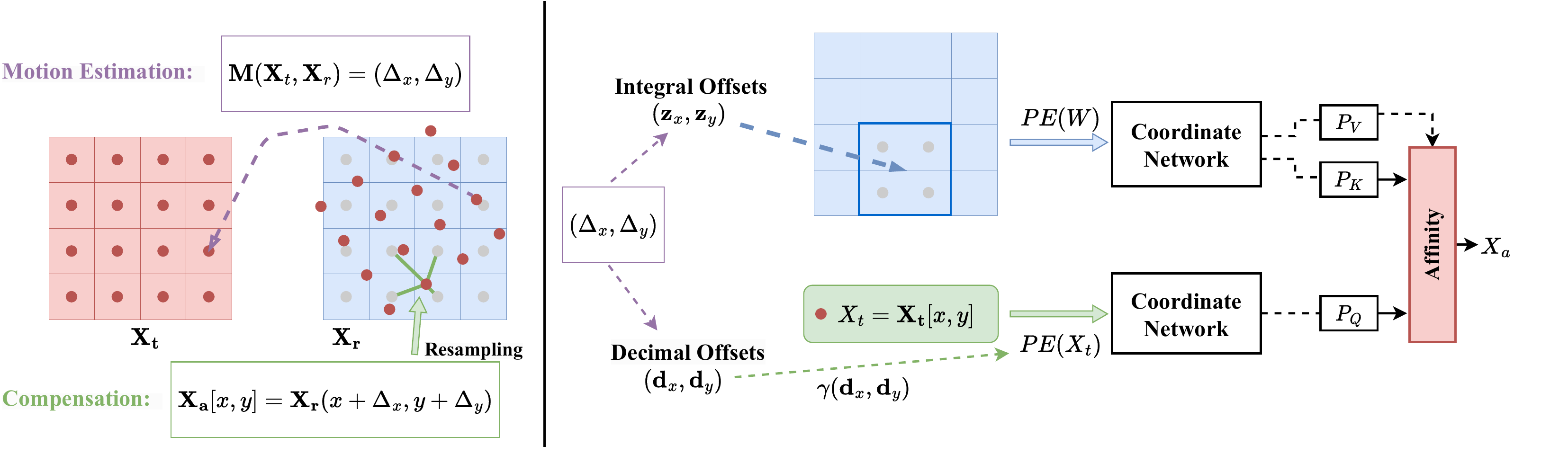}%
\caption{(a). Motion estimation provides a transformation that maps the reference frame $\referenceX$ to the current frame $\currentX$. Compensation performs resampling on $\referenceX$ to obtain the aligned value $\alignedX[x,y]$ at each pixel location. (b) The estimated motion offsets are decomposed into integral offsets and decimal offsets. The integral offsets are used for window queries and the decimal offsets are used for position encoding for the query pixel $X_t$. The features along with the positional encodings are modeled with coordinate networks, and the aligned pixel $X_a$ is obtained by a cross-attention mechanism.}\label{fig:preliminary_method}
\end{figure*}
 
\section{Preliminaries}\label{sec:preliminary}

\subsection{Spatial Alignment}
In video super-resolution, inter-frame propagation enhances information across time.  The propagation is facilitated by spatial alignment; the aligned frame gets concatenated with the current frame, and the two are fed together into subsequent network blocks.  The alignment can be performed on either input images or intermediate feature maps. We refer to both as frame-wise alignment within a general formulation and denote both images and features simply as some $\mathbf{X} \in \mathbb{R}^{H \times W \times C}$ where $H$, $W$ and $C$ are the height, width and channels, respectively.  To focus on the spatial operations in the subsequent discussion, we omit the $C$ dimension and mention it explicitly only where needed.  

As shown in Fig.~\ref{fig:preliminary_method}a, alignment can be broken down into two steps:  (1) motion estimation 
and  (2) motion compensation.  
Basic implementations of alignment perform these two steps in a one-off manner~\cite{chan2021basicvsr}. More advanced methods make multiple motion estimates and ensemble multiple compensations with convolution, through deformable convolution \cite{wang2019edvr,chan2022basicvsrpp}, or with attention mechanisms, through deformable attention \cite{liang2022recurrent_rvrt}.  

Consider a current frame $\currentX$ indexed by $t$ with spatial coordinates $[x,y]$ \footnote{In our work, we will use square and round braces to emphasize the difference between between discrete coordinates on a pixel grid versus continuous coordinates on a continuous plane.}, and a reference frame $\referenceX$ indexed by $r$. 
Corresponding points in $\currentX$ and $\referenceX$ are related by a motion displacement field $\motion \in \mathbb{R}^{H \times W \times 2}$. Each element $\motion[x,y]=(\Delta_x, \Delta_y)$ represents the displacement of the pixel at coordinate $[x,y]$ in $\currentX$ to its corresponding point in the reference frame $\referenceX$, with coordinates in the reference given by $(x+\Delta_x, y+\Delta_y)$. A simple way to estimate $\motion$ is by solving for the optical flow between $\currentX$ and $\referenceX$. More recent works~\cite{wang2019edvr,chan2022basicvsrpp,liang2022recurrent_rvrt} estimate additional offsets to refine the predicted optical flow.

Based on $\motion$, frame-wise alignment estimates $\alignedX \in \mathbb{R}^{H \times W \times C}$, which can be regarded as a motion-compensated version of the reference frame $\referenceX$ :
\vspace{-0.5em}
\begin{equation}\label{eq:alignment}
    \alignedX = \mathcal{W}(\referenceX, \motion),
\end{equation}
where $\mathcal{W}$ indicates a warping function that performs the motion compensation. 
The standard strategy for compensation is through backward warping, where the following estimation is iterated on all spatial locations for $\alignedX$: 
\begin{equation} \label{eq:backwardwarp}
\alignedX[x,y]= \referenceX(x+\Delta_x,y+\Delta_y).
\end{equation}
\noindent Note that estimating $\referenceX(x+\Delta_x,y+\Delta_y)$ requires a resampling operation on $\referenceX$, as $(\Delta_x,\Delta_y)$ are continuous values.

\subsection{Spatial Resampling for Alignment}\label{sec:resample}
Spatial resampling estimates sub-pixel values on a discrete image or feature grid $\mathbf{X}$. The \emph{support} of a resampling method indicates the window on $\mathbf{X}$ which is required to estimate the value $\mathbf{X}(a,b)$ for continuous coordinates $(a,b)$.  Common methods for spatial resampling interpolate from the support on a heuristic basis.  Examples include nearest-neighbour, bilinear and bicubic interpolation.

\noindent\textbf{Nearest-neighbour Interpolation} The resampled value is taken as the value of $\mathbf{X}$ at the discrete coordinates $[x,y]$ nearest to $(a,b)$:
\begin{align}
\mathbf{X}(a,b)_{\text{nn}} & = \mathbf{X}[x^*,y^*],\\
  \text{where } [x^*,y^*]  & = \argmin_{(x,y)} || (a,b) - (x,y) ||_2.
\end{align}

\noindent\textbf{Bilinear / Bicubic Interpolation} estimates the resampled value as a weighted sum of the 4 (bilinear) or 16 (bicubic) discrete neighbours around $(a,b)$, which we denote in short form as $\lfloor a,b \rceil$: 
\begin{equation}
\referenceX(a,b)_{\text{bi}} = 
\sum_{(x,y) \in \lfloor a, b \rceil_{bi}} w_{xy}\cdot \mathbf{X}[x,y],\\
\end{equation}

where $w_{xy}$ are the associated weighting coefficients based on either a linear (bilinear) or quadratic (bicubic) interpolation of $a$ and $b$ with respect to the neighbouring coordinates\footnote{Strictly speaking, the interpolation is only linear (or quadratic) for along lines parallel in the $x$ and $y$ directions, \ie at fixed points. }. We refer the reader to Sec. A of the Supplementary for the precise definitions.

Without prior knowledge on how the original discrete image $\referenceX$ is sampled, most interpolation methods impose smoothness assumptions for resampling. %
By virtue of assuming linear or quadratic interpolants, bilinear and bicubic interpolation enforce an L0 / L1 smoothness constraint on the underlying image plane. %
Such constraints are equivalent to applying low-pass filters on the source frame's intensity or features~\cite{youssef1998analysis}, hence the blurry interpolated results. 
Notably, nearest-neighbour interpolation does not have any smoothness requirements, and hence does not have a low-pass effect.  However, it introduces spatial distortions by shifting the sampled position to the nearest pixel grid.

\subsection{Analysis on Resampling for Alignment}

We examine the frequency response of the nearest and bilinear interpolation methods. Let $f_s$ denote the sampling frequency. The nearest-neighbour interpolator corresponds to a rectangular function in the spatial domain and its Fourier transform is a sinc function given by $F_{nn}(f) = \mathop{\rm sinc}(f/f_s) $, which has a decay rate of $f_s/f$ in the out-of-band region. The bilinear interpolator corresponds to a triangular function in the spatial domain and its Fourier transform is a squared sinc function given by $F_{bi}(f)= \mathop{\rm sinc}^2(f/f_s) $, which has a decay rate of $(f_s/f)^2$ in the out-of-band region.

Compared to the nearest-neighbour interpolator, which has a decay rate of $f_s/f$, the bilinear interpolator with a decay rate of $(f_s/f)^2$ can suppress more out-of-band aliasing artifacts. This explains why the nearest-neighbour interpolator introduces more artifacts and distortion. However, the bilinear interpolator also causes more smoothing effect on the in-band frequency than the nearest-neighbour interpolator. In the following section, we investigate the use of coordinate networks as function approximators for the ideal interpolator.

\section{Methodology}
\subsection{Coordinate Network for Implicit Resampling}

A coordinate network is a network that uses coordinates as inputs to represent signals. We use a coordinate network as a prior for resampling and encode the prior as trainable weights in an PE-MLP. Such a use of  coordinate networks was first explored in neural priors for scene flow regularization~\cite{li2021neural_scene_flow_prior} though an implicit optimization at runtime.

During training, the coordinate network is jointly optimized with an L2 loss on all alignment instances. Being a universal approximator in theory \cite{hornik1989multilayer}, MLPs can represent any function and frequency. Moreover, we use positional encoding(PE)-MLPs, as they have been shown to have good learning capacity for high frequency content.~\cite{mildenhall2021nerf} %

Specifically, given the input feature $\mathbf{X}$ and its coordinates $\vp$, the coordinate network $F$ jointly modelling feature and its position.
\begin{equation}
    \mathbf{R} = F(\mathbf{X} + \gamma(\vp)) 
\end{equation}
\noindent where $\gamma(\vp)$ denotes a positional encoding and $R$ is the output feature.  
The positional encoding $\gamma(\vp) \in \mathbb{R}^2 \rightarrow \mathbb{R}^{4D} $ is computed by projecting low-dimensional input coordinates $\vp$ to a $4D$ dimensional hypersphere. 
\begin{equation}
    \!\gamma(\vp)\! =\! \big[[\sin(\omega\vp), \cos(\omega\vp)],\dots,[\sin(\omega^{D\!-\!1}\vp), \cos(\omega^{D\!-\!1}\vp)]\big],
\end{equation}
\noindent where $\omega$ is the angular speed and $D$ controls the number of frequency bands from $\omega$ to $\omega^{D\!-\!1}$.  A larger $D$ provides higher capacity for encoding higher frequency.

Coordinate networks offer several advantages over conventional alignment methods. First, they can theoretically represent any frequency component of the signal, thus avoiding the low-pass filtering effect. Second, they can serve as a general alignment prior that can be applied to any alignment scenario, regardless of the feature scale or the alignment configuration. In contrast, existing alignment modules are usually tailored for specific feature scales and alignment configurations, which may limit their generalization ability and increase their parameter size.

\subsection{Alignment with Implicit Resampling}

{Having obtained the output feature through the coordinate network, we conducting spatial alignment via a cross-attention mechanism.} Our key insight is that in spatial alignment, the values of the current frame $\currentX$ can also benefit the compensation.  In conventional methods, including deformable convolution and deformable attention, the support for the compensation is based only on the values of the reference frame $\referenceX$. The values of the current frame $\currentX$, beyond estimating the displacement field $\motion$, are not used. In contrast, we use as support values from both $\referenceX$ and $\currentX$, which we find can help us \tp{improve the alignment accuracy}.  To that end, we propose an alignment where the resampling is \emph{implicit}.  Rather than estimate the resampled value with an explicit function, as the examples given in Sec.~\ref{sec:resample}, we \tp{align} with a {cross-attention operation between the the corresponding outputs from reference and current frames, where the $\currentX$ serves as query, $\referenceX$ as key and the values.}

\subsection{Window-based Cross Attention}\label{sec:wbattention}

We define the motion-compensation for coordinate $[x,y]$:

\begin{equation}\label{eq:iw}
\alignedX[x,y] = \text{softmax}\Big(\frac{\mathbf{Q} \mathbf{K}^T}{\sqrt{C}}\Big)\mathbf{V}
\end{equation}
where 
\begin{align}
\mathbf{Q} &= F_{q}(X_t+P_t), \\
\mathbf{K} &= F_{k}(\mathbf{W}_r+\mathbf{P}_r), \\
\mathbf{V} &= F_{v}(\mathbf{W}_r+\mathbf{P}_r)    
\end{align}
are the corresponding output from the coordinate networks; $\text{softmax}\Big(\frac{\mathbf{Q} \mathbf{K}^T}{\sqrt{C}}\Big)$ is the affinity matrix encoding the similarity of pixel  $\currentXpixel \in \mathbb{R}^{1 \times C}$ from the current frame and a window of pixels $ \mathbf{W}_r \in \mathbb{R}^{w \times w \times C}$ from the reference frame. 

The window center is based on the estimated displacement, where $w$ is the chosen window size. Specifically, for $\motion(x,y) = (\Delta_x, \Delta_y)$, we can split it into integer part $(\mathbf{z}_x, \mathbf{z}_y)$ and decimal part $(\mathbf{d}_x, \mathbf{d}_y)$:
\begin{equation}
    (\Delta_x, \Delta_y)=(\mathbf{z}_x, \mathbf{z}_y)+(\mathbf{d}_x, \mathbf{d}_y).
\end{equation}
The integer part selects the window of support in $\referenceX$, while decimal part is encoded into a positional encoding to estimate the sub-pixel information from the window of support. 
The sub-pixel information is then used to encode coordinate information for $X_t$ and $X_a$, as it reflects the relative position between queried pixel and neighbouring pixels.

\noindent\textbf{Integral Offsets as Window Queries} The window $\mathbf{W}_r$ is centered on $(x+\mathbf{z}_x,y+ \mathbf{z}_y)$ and selects the neighbouring $w\times w$ pixels, where
\begin{align}
\mathbf{W}_r[i, j]&=\referenceX[ x\!+\!\mathbf{z}_x\!+\!i,\  y\!+\!\mathbf{z}_y\!+\!j] \\
\mathbf{P}_r[i, j]&=\gamma([i,j]/w)
\\ &\forall -\lfloor w/2\rfloor \leq i,j \leq w \!-\! \lfloor w/2 \rfloor\! -\!1 . 
\end{align}

For window pixels, the positional encoding is given as a normalized relative position to the window center, hence the scaling by $1/w$.

{The window-based attention} %
reduces the computational cost to $O(w^2\!\cdot\!HW)$ from the quadratic cost $O(HW\!\cdot\!HW)$ of the global attention. The choice in window size $w$ is flexible for \tp{different motion accuracies}. Generally, larger $w$ is more robust to noisy motion estimation while smaller $w$ provides sharper results. 

\noindent\textbf{Decimal Offsets as Positional Encoding} For the query encoding for pixel $[i,j]$, we have
\begin{align}
\currentXpixel & = \currentX[x,y] \\
P_t&=\gamma([\mathbf{d}_x, \mathbf{d}_y]/2w),
\end{align}

\noindent where the positional encoding is again normalized with respect to the window center. 
As $[\mathbf{d}_x, \mathbf{d}_y]/w$ is a decimal, a high angular speed $\omega$ is required to represent this information. 
For 
\begin{equation}
    \omega = {T^{-D}},
\end{equation}

\noindent we set $T=0.01$ and form a geometric progression from $2\pi$ to $100\pi$ on the angular speed to represent more precise sub-pixel position information.

\begin{figure*}
\begin{minipage}{0.55\textwidth}
     
    \includegraphics[width=0.5\linewidth]{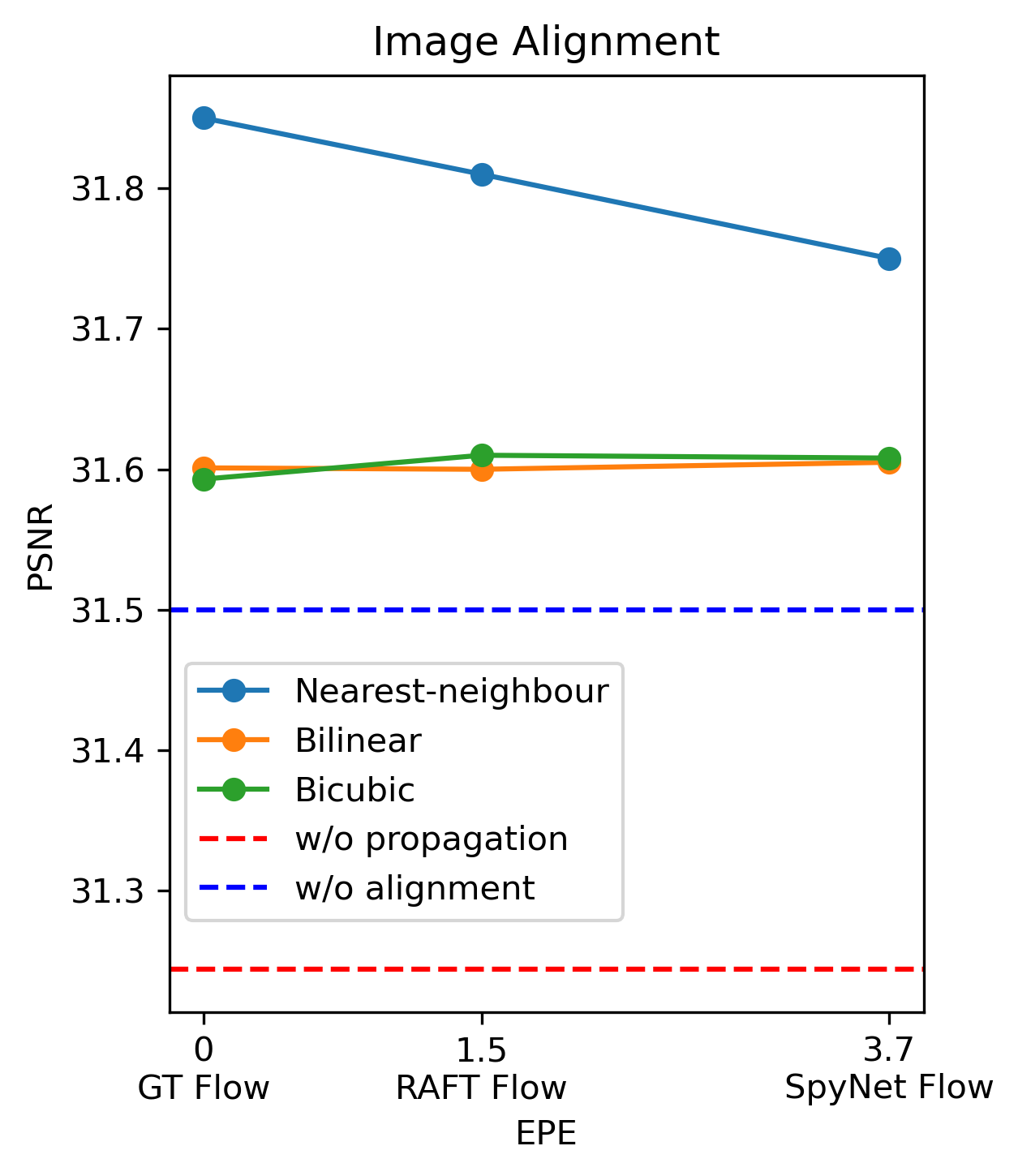}
    \hspace{-0.5em}
    \includegraphics[width=0.5\textwidth]{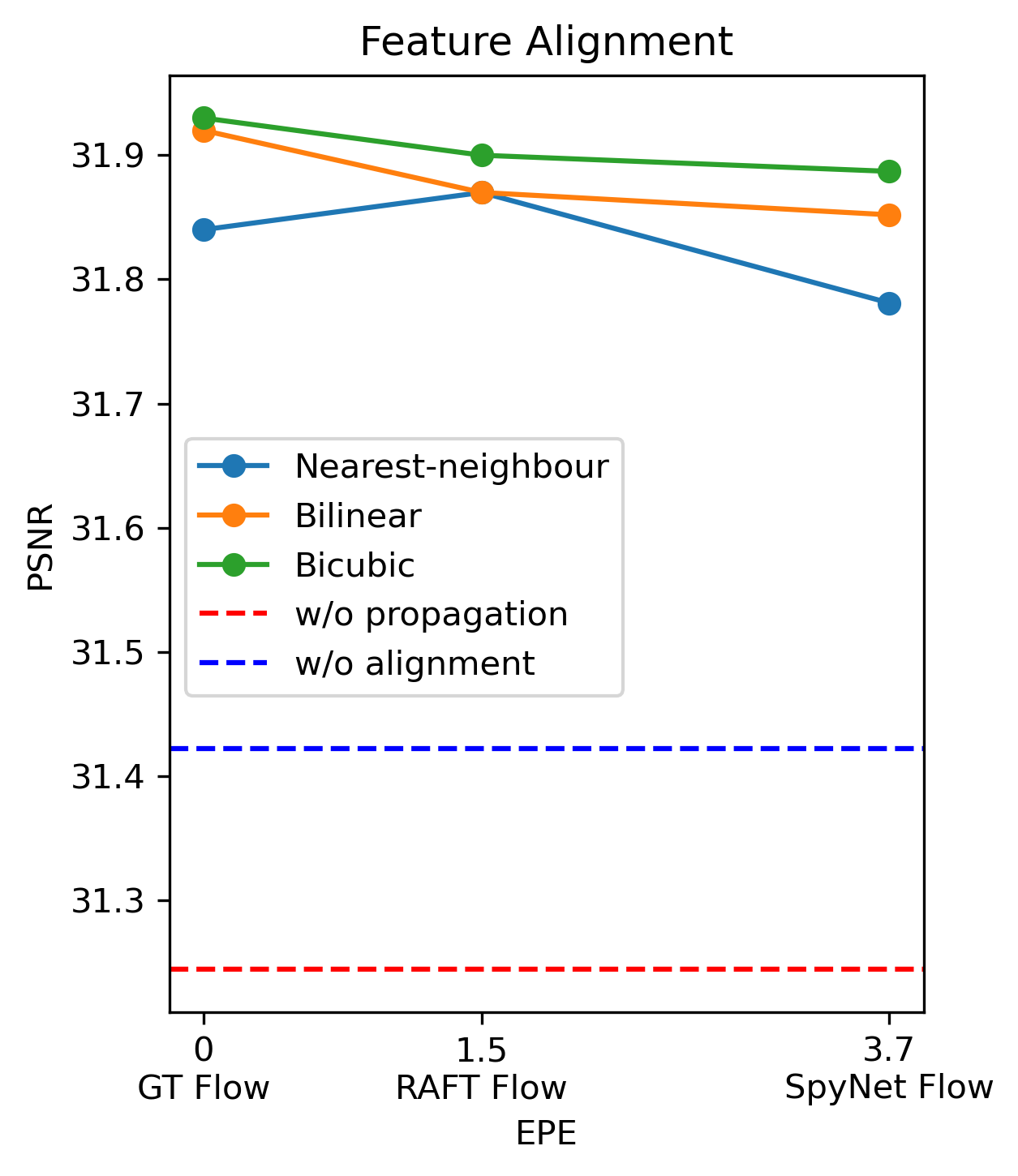}
    
\end{minipage}
  \begin{minipage}{0.43\textwidth}
    \centering

\resizebox{\textwidth}{!}{%

\vspace{7em}
\begin{tabular}{l|c|c|c|c|c}
\hline
Alignment & Params & Resamp. &GT & RAFT        & SpyNet\\
            & (M)   &       &   Flow    &   Flow    & Flow\\
\hline
\multirow{3}{*}{OF Warp}        &\multirow{3}{*}{1.35}
& nearest.  & 31.84 & 31.87 & 31.78  \\
&& bilinear & 31.92 & 31.87 & 31.85  \\
&& bicubic  & 31.93 & 31.90 & 31.89  \\

\hline
FGDC \cite{wang2019edvr}        &1.60& bilinear &  32.08  &  31.99  &  31.98  \\
FGDA \cite{liang2022recurrent_rvrt}  &1.56& bilinear &  32.03  &  31.91  &  31.94  \\
PA \cite{shi2022rethinking}     &1.35& nearest. &  31.81  & 31.85  &  31.82  \\
IA (ours)                      &1.36&  implicit & \textbf{32.14}  & \textbf{32.03}  & \textbf{32.05}  \\
\hline
\end{tabular}
}
\captionof{table}{Comparisons on feature alignment. Implicit Resampling-based Alignment (IA) outperforms all three state-of-the-art alignment methods.}\label{tab:alignment_sota}
\end{minipage}
  
  \caption{Comparison of PSNR on alignment utilizing optical flow with different accuracies. Nearest-neighbour interpolation outperforms bilinear interpolation for image alignment, while the opposite is true for feature alignment. This observation highlights inherent limitations associated with both interpolation techniques.}
  \vspace{-1.5em}
  \label{fig:alignment_interpolations}
\end{figure*}

\section{Experiments on Resampling for Alignment}

We perform alignment studies under a synthetic dataset with ground truth optical flow, as well as two commonly used optical flows in video super-resolution, namely RAFT \cite{teed2020raft} and SPyNet \cite{ranjan2017optical_spynet}. The former is a precise and slow method, while the latter is faster but less accurate.

For the synthetic data, we split the training videos of the clean data track of Sintel~\cite{butler2012naturalistic_sintel} into 20 training and 3 testing videos and report the testing results. We generate low-resolution training pairs with bicubic down-sampling of the high-resolution counterparts. As only first-order forward backward optical flow ($t \! \rightarrow \! t\!+1$) ground-truth is provided, we perform image alignment from $(t\!+\!1 \! \rightarrow \! t)$ and concatenate with original frame before feeding into the super-resolution network. 

We use a VSR transformer~\cite{liang2022vrt_vrt} as the super-resolution backbone. 
We consider the following baselines and alignment strategies: (1) w/o Prop.: An image super-resolution baseline with no propagation. (2) w/o Align.: propagation without alignment. (3) Optical flow warping with nearest-neighbour , bilinear and bicubic interpolations. (4) Flow-Guided Deformable Convolution (FGDC)~\cite{wang2019edvr}. (5) Flow-Guided Deformable Attention (FGDA)~\cite{liang2022recurrent_rvrt} (6) Patch alignment (PA)~\cite{shi2022rethinking}. (7) Our implicit resampling-based Alignment (IA).

\subsection{Results Analysis} 

\noindent\textbf{The Impact of Resampling}  
Fig.~\ref{fig:alignment_interpolations} compares PSNR values across various alignment methods. Intriguingly, nearest-neighbour interpolation outperforms bilinear interpolation for image alignment, while the opposite is true for feature alignment. This observation highlights inherent limitations associated with both interpolation techniques. Specifically, nearest-neighbour introduces distortions, whereas bilinear interpolation techniques introduce smoothing effects. Our conclusion is grounded in two primary observations. 

Firstly, for image alignment, the frames have relatively high frequency components as it has not passed through any convolution layers (which themselves act as smoothing filters). As such, any spatial distortions introduced by nearest-neighbour interpolation are outweighed by its ability to preserve high-frequency components.

Secondly, for feature alignment, there is reduced sensitivity to high-frequency components because the features are likely concentrated in lower frequency spectrums due to the spectral bias of neural networks~\cite{rahaman2019spectral, chen2021ssd}. \tp{Thus the gains from preserving high-frequency are outweighed by the introduced spatial distortions for nearest-neighbour interpolation. }
In light of these observations, we posit that an optimal resampling method should not impose smoothness constraints to avoid attenuating high-frequency components \emph{and} mitigate the distortions resulting from coordinate quantization.

\noindent\textbf{Comparison with State-of-the-Art Alignment Methods}
\tp{Given the established effectiveness of feature alignment over image alignment, our comparison focuses solely on state-of-the-art approaches in feature alignment.} From Tab. ~\ref{tab:alignment_sota}, Implicit Resampling-based Alignment (IA) outperforms all three state-of-the-art alignment methods, owing to its capacity to implicitly learn resampling weights. In contrast, FGDC, FGDA, and PA rely on adaptations of bilinear and nearest-neighbour interpolation, introducing either smoothing priors or distortions, contributing to their comparative performance inferiority. \tp{FGDA is inferior to FGDC due to the limited training data.
As PA is a robust method designed to counter inaccurate optical flow. The synthetic dataset with GT flow is not the ideal case for PA so it doesn't do well.}
Regarding parameter size considerations, 
IA, functioning as a coordinate network, shares parameters across all alignment operations, resulting in a modest parameter increase of 0.01M compared to FGDC (0.25M) and FGDA (0.21M).

\section{Comparison with State-of-the-Art Methods on Large-Scale Datasets}

\begin{table*}[t!]
  \small
  \centering

  \label{tab:BI}
  \resizebox{0.80\textwidth}{!}{
  \begin{tabular}{l|c|c|c|cc|cc|cc}
    \hline
    \multirow{2}{*}{Method}& \multirow{2}{*}{Propagation}&  Frames & Params & \multicolumn{2}{c|}{REDS4} & \multicolumn{2}{c|}{Vimeo-90K-T} & \multicolumn{2}{c}{Vid4}\\
    
    & &REDS/Vimeo & (M) & PSNR & SSIM & PSNR & SSIM & PSNR & SSIM \\
    \hline
    TOFlow~\cite{xue2019video_tof}  & \multirow{5}{*}{First Order}  &       5/7 & - & 27.98 & 0.7990  & 33.08 & 0.9054  & 25.89 & 0.7651\\
    EDVR~\cite{wang2019edvr}      &  & 5/7 & 20.6 & 31.09 & 0.8800  & 37.61 & 0.9489 & 27.35 & 0.8264 \\
    MuCAN~\cite{li2020mucan}      &  &     5/7  & - & 30.88 & 0.8750  & 37.32 & 0.9465 &  - & - \\

    BasicVSR~\cite{chan2021basicvsr}& &15/14 & 6.3 & 31.42 & 0.8909 & 37.18 & 0.9450 & 27.24 & 0.8251\\
    IA-CNN (ours) & &15/14 & 8.5 & \textbf{31.68} & \textbf{0.8959} & \textbf{37.34} & \textbf{0.9463} & \textbf{27.42} & \textbf{0.8315}\\
    \hline
    BasicVSR++~\cite{chan2022basicvsrpp}& \multirow{5}{*}{Second Order} &30/14 & 7.3  & 32.39 & 0.9069 &  37.79 & 0.9500 & 27.79 & 0.8400\\
    VRT  &  &16/7 & 35.6 & 32.19 & 0.9006 & {38.20} & {0.9530} & 27.93 & 0.8425\\
    RVRT \cite{liang2022recurrent_rvrt} & & 30/14 &  10.8 & {32.75} & {0.9113} & 38.15 & 0.9527 & {27.99} & {0.8462} \\
    PSRT-recurrent~\cite{shi2022rethinking}  & & 16/14 & 13.4 & {32.72} & {0.9106} & \textbf{38.27} & \textbf{0.9536} & {28.07}& {0.8485}\\
    IA-RT (ours) &  &16/7 & 13.4 & \textbf{32.90} & \textbf{0.9138} & 38.14 & 0.9528 & \textbf{28.26} & \textbf{0.8517}\\
    \hline

  \end{tabular}}
    
    \vspace{-0.5em}\caption{Quantitative comparison on REDS4~\cite{nah2019ntire}, Vimeo-90K-T~\cite{xue2019video_tof} and Vid4~\cite{liu2013bayesian} dataset for $4\times$ Video SR.}  
    \label{tab:comparision_sota}
    
\end{table*}

On standard video SR datasets REDS~\cite{nah2019ntire}, Vimeo90K~\cite{xue2019video_tof} and Vid4~\cite{liu2013bayesian}, we incorporate implicit resampling-based alignment into two state-of-the-art networks: a convolutional neural network (CNN) based model (BasicVSR~\cite{chan2021basicvsr}) for first-order VSR, which leverages information from one neighboring frame, and a recurrent Transformer based model (PSRT-recurrent~\cite{shi2022rethinking}) for second-order VSR, which utilizes information from two neighboring frames. We denote our models as IA-CNN and IA-RT, respectively. We refer the reader to Sec. B of the Supplementary for exact experimental configurations.

\begin{figure*}[h!]
\centering

\includegraphics[width=\textwidth]{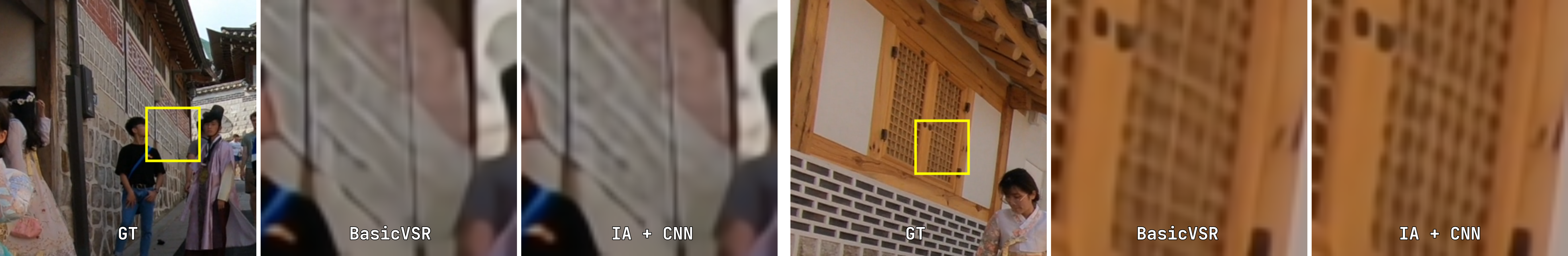}

\caption{Qualitative comparisons on REDS4 dataset. IA-CNN provides more details on the wall and more uniform patterns on the window. }\label{fig:qual_reds4}
\vspace{-1em}
\end{figure*}

\begin{figure*}[ht!]
\centering
\includegraphics[width=\textwidth]{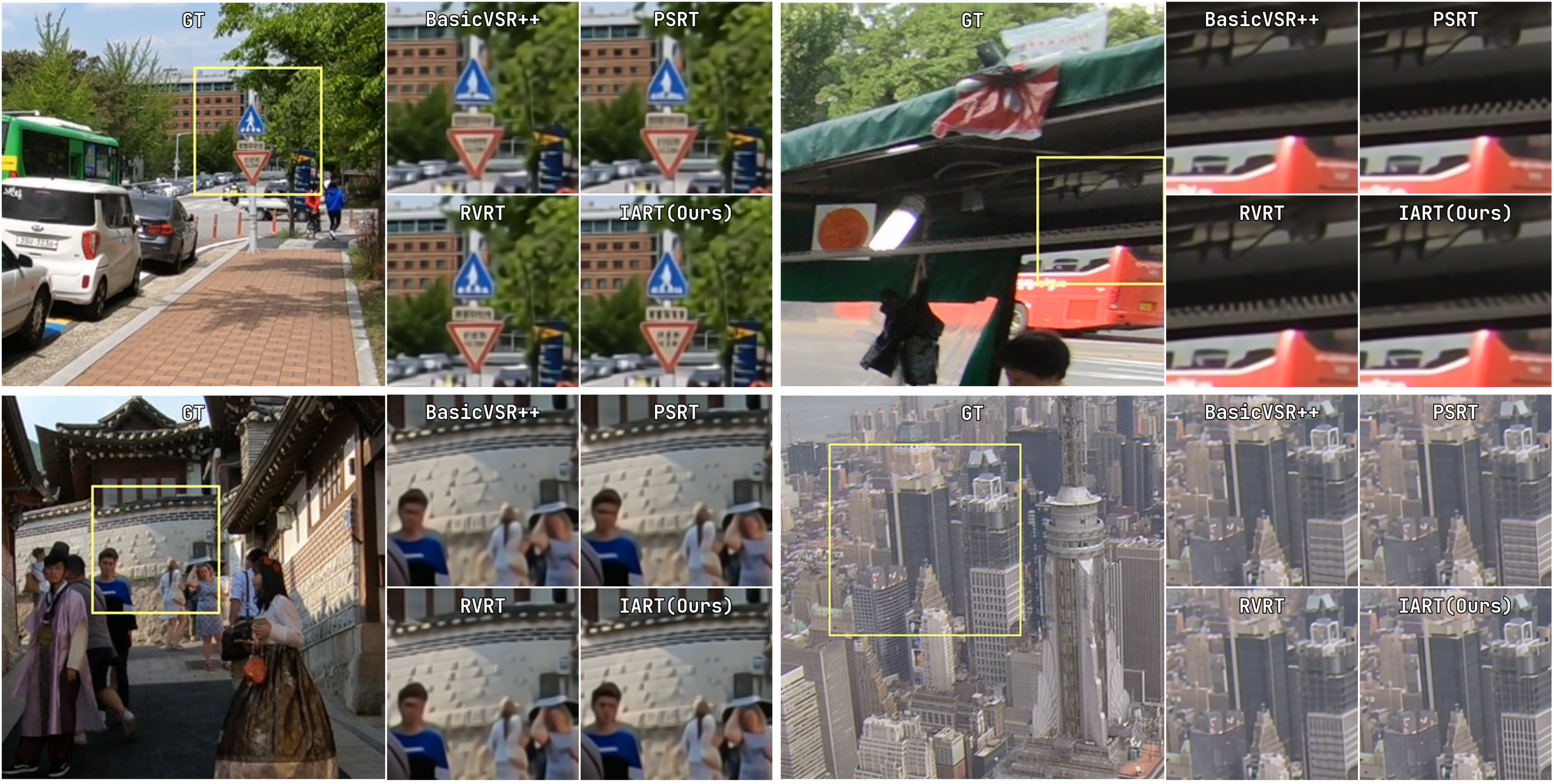}
\vspace{-1em}
\caption{Qualitative comparisons on REDS4 and Vid4. IA-RT provides sharper results and more fine-grained patterns.}\label{fig:qual_iart}
\vspace{-1em}
\end{figure*}

\begin{figure*}[t!]
\centering

\includegraphics[width=0.95\textwidth]{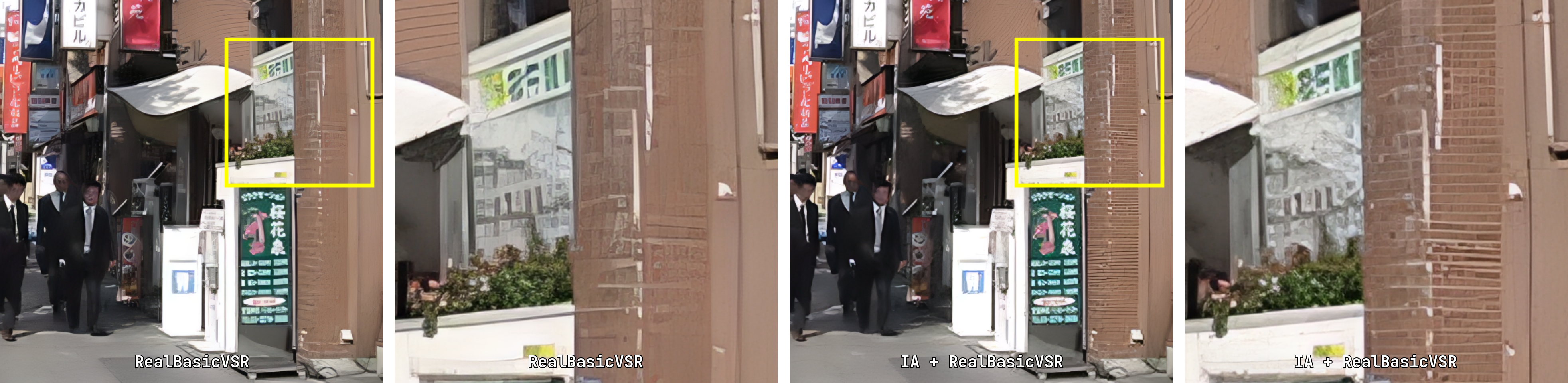}
\\
\includegraphics[width=0.95\textwidth]{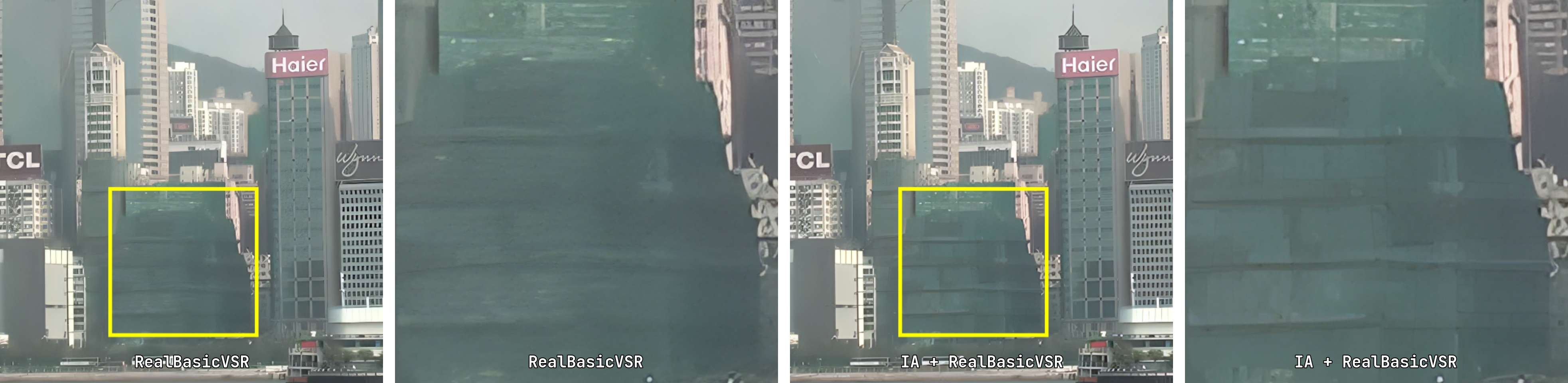}%
\caption{Qualitative comparison on VideoLQ dataset. Our proposed IA method recovers the brick textures and the wall patterns, which RealBasicVSR does not recover. We highlight the detail regions with yellow boxes.}\label{fig:qual_realvsr}
\end{figure*}

\subsection{Results Analysis}
Table \ref{tab:comparision_sota} presents a quantitative comparison with state-of-the-art (SOTA) methods. 
For CNN-based models, IA-CNN outperforms its baseline BasicVSR with only are marginal increase in parameters. 
For Transformer-based models, IA-RT achieves the highest accuracy in the configuration with 6 training input frames. It outperforms its baseline, PSRT-recurrent, by $0.27\!\uparrow$ on PSNR and $0.0046\!\uparrow$ on SSIM. In the configuration with 16/7 input frames, IA-RT surpasses PSRT-recurrent by $0.18\!\uparrow$ on PSNR and $0.0032\!\uparrow$ on SSIM for REDS4, $0.19\!\uparrow$ on PSNR and $0.0032\!\uparrow$ on SSIM for Vid4, establishing itself as the current state-of-the-art on these two datasets. Yet our implicit alignment module only introduce 0.2\% parameters compare to its baseline ~\cite{shi2022rethinking}.
IA-RT is slightly below PSRT on Vimeo90k, primarily due to challenges in accurately estimating optical flow, limiting the benefits of accurately sampling at a sub-pixel level.

\noindent\textbf{Qualitative Results for IA-CNN and IA-RT}
Fig.~\ref{fig:qual_reds4} shows qualitative comparisons between BasicVSR and IA-CNN on the REDS4 dataset. Fig.~\ref{fig:qual_iart} shows qualitative comparisons between BasicVSR++, PSRT, RVRT and IART. The IA-CNN and IART exhibit enhanced ability to propagate high-frequency contents and reconstruct finer patterns compared to other methods. Additional qualitative results can be found in the Sec. C of the Supplementary.

\noindent\textbf{Real-World Video SR} is a variant of the video SR task where the low-resolution inputs are corrupted with non-deterministic degradation
such as blur, noise, and compression artifacts. In the face of such degradation, 
existing methods often yield excessively smoothed results. 
Qualitative comparisons in \cref{fig:qual_realvsr} showcase that, when integrated into RealBasicVSR~\cite{chan2022investigating}, our implicit alignment method produces results that are more realistic and fine-grained. 

\subsection{Ablation Studies}

\noindent\textbf{Positional Encoding}
Table \ref{tab:pe} 
shows that having positional encoding yields a noteworthy improvement in PSNR by 0.28 compared to the naive window-based cross-attention. 
\tp{When positional encodings are only enabled for window indices, a large drop on PSNR is observed, suggesting the scenario where the estimated motion is quantized to integers will leads to degraded results.}
When only introducing positional encodings on decimal offsets, the model collapse. This outcome is attributed to the absence of relative positional information for the window features. 
\vspace{-0.5em}

\begin{table}[h]
\centering
\resizebox{0.95\linewidth}{!}{%
\begin{tabular}{c|c|c|c}
\hline
 PE on decimal offsets & PE on window indices & PSNR & SSIM   \\
\hline
\xmark      & \xmark    & 30.43 & 0.8700   \\
\cmark      & \xmark    & 28.71 & 0.8184   \\
\xmark      & \cmark    & 30.54 & 0.8730   \\
\cmark      & \cmark    & 30.71 & 0.8776   \\
\hline
\end{tabular}
}
\vspace{-0.5em}

\caption{Ablations on positional encodings.
}\label{tab:pe}
\vspace{-1em}
\end{table}

\noindent\textbf{Window Size}
The PSNR/SSIM results corresponding to different window sizes for the cross-attention operation are presented in \cref{tab:window_size} on Sintel. Larger window sizes result in a more extensive receptive field, but concurrently diminish alignment quality due to increased noise. 
However, a larger window size proves advantageous, contributing to increased model robustness in the context of Real-world Video Super-Resolution (VSR), where predicting accurate optical flow poses challenges.

\begin{table}[h]
\centering
\footnotesize
\begin{tabular}{c|c|c|c}
\hline
Window Size & 2x2 &3x3 & 4x4    \\
\hline
GT Flow & 32.06/0.9024  &32.06/0.9021 &32.05/0.9019        \\
SpyNet Flow & 31.97/0.9004  &31.95/0.9005 &31.96/0.9005        \\
\hline
\end{tabular}
\vspace{-1em}
\caption{Ablations on different window sizes for GT flow and SpyNet flow on Sintel dataset.}\label{tab:window_size}
\vspace{-1em}
\end{table}

\subsection{FLOPS and Runtime Comparison}

\cref{tab:flops} gives he comparison of parameters, FLOPs, and runtimes for IART and other VSR model. We re-estimate the inference time for both PSRT-recurrent on RTX-A5000.

\begin{table}[t]
\footnotesize
    \centering
    \begin{tabular}{l|c|c|c}
    \hline
        Method & Param. (M) & FLOPs (T) & Runtime (ms) \\
    \hline
        EDVR \cite{wang2019edvr} & 20.6 & 2.95 & - \\
        VSRT \cite{cao2021video} & 32.6 & 1.60 & -\\
        VRT \cite{liang2022vrt_vrt} & 35.6 & 1.30 & -\\
        PSRT-recurrent \cite{shi2022rethinking} & 13.4 & 1.50 & 2020$^\dagger$ \\
        IART (ours) & 13.4 & 1.62 & 2105\\
    \hline
    \end{tabular}
    \vspace{-1em}
    \caption{
The comparison of parameters, FLOPs, and runtimes. 
}    \label{tab:flops}
\vspace{-1em}

\end{table}

\section{Conclusion}\label{sec:conclusion}

This paper investigates the impact of resampling on alignment for video super-resolution through experiments conducted on a synthetic dataset employing ground-truth optical flow. Our findings underscore the necessity for resampling techniques to preserve the original sharpness of features and avoid distortions for effective alignment. We propose an implicit resampling-based alignment method using coordinate networks and window-based cross-attention, by incorporating estimated motions encoded into positional encoding. Our proposed method exhibits superior performance compared to state-of-the-art alignment techniques on both synthetic and real-world datasets.

An inherent drawback of implicit resampling-based alignment is the reduced interpretability of the alignment process. However, we argue that it can be validated through extensive testing and experimentation.

{
    \small
    \bibliographystyle{ieeenat_fullname}
    \bibliography{main}
}

\end{document}


\title{Enhancing Video Super-Resolution via Implicit Resampling-based Alignment}

\author{
Kai Xu\quad Ziwei Yu \quad  Xin Wang\quad Michael Bi Mi\quad Angela Yao
\\\vspace{1em} \texttt{kxu@comp.nus.edu.sg}
}

\maketitle
\section{Methods}
\subsection{Bilinear / Bicubic Interpolation} Bilinear / Bicubic Interpolation estimates the resampled value as a weighted sum of the 4 (bilinear) or 16 (bicubic) discrete neighbours around $(a,b)$, which we denote in short form as $\lfloor a,b \rceil$: 
\begin{align}
 \referenceX(a,b) = 
\sum_{(x,y) \in \lfloor a, b \rceil_{bi}} w_{xy}\cdot \mathbf{X}[x,y], 
\end{align}

where $w_{xy}$ are the associated weighting coefficients based on either a linear (bilinear) or quadratic (bicubic) interpolation of $a$ and $b$ with respect to the neighbouring coordinates. 

For bilinear interpolation:
\begin{equation}
    w_{xy}  = \tp{|a-x|*|b-y|} 
\end{equation}

For bicubic interpolation, please refer to~\cite{keys1981cubic} for detailed definition.

\subsection{Network Structures}

We employed a single-layer fully connected layer as the encoder for $F_q$, $F_k$, and $F_v$ in our experiments. These experiments were conducted on both CNN-based backbones, following the architecture of BasicVSR~\cite{chan2021basicvsr}, and Transformer-based backbones, following the PSRT-recurrent model \cite{shi2022rethinking}. Alignment was applied for either first-order or second-order bidirectional propagation, with a default window size of 2x2 unless explicitly stated.

\begin{figure*}
\centering
\includegraphics[width=\textwidth]{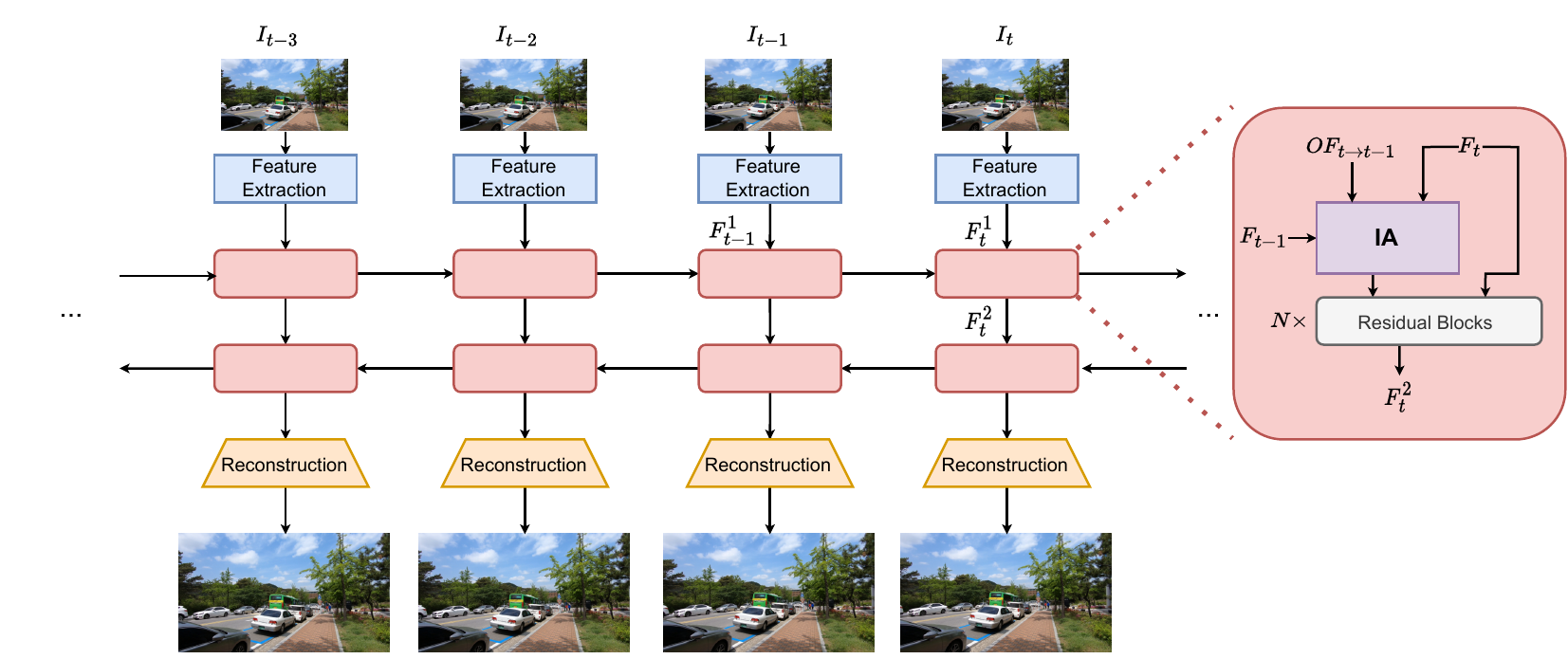}%

\caption{Network Structure for IA-CNN.}\label{fig:sup_network_structures_iacnn}

\end{figure*}

\begin{figure*}
\centering
\includegraphics[width=\textwidth]{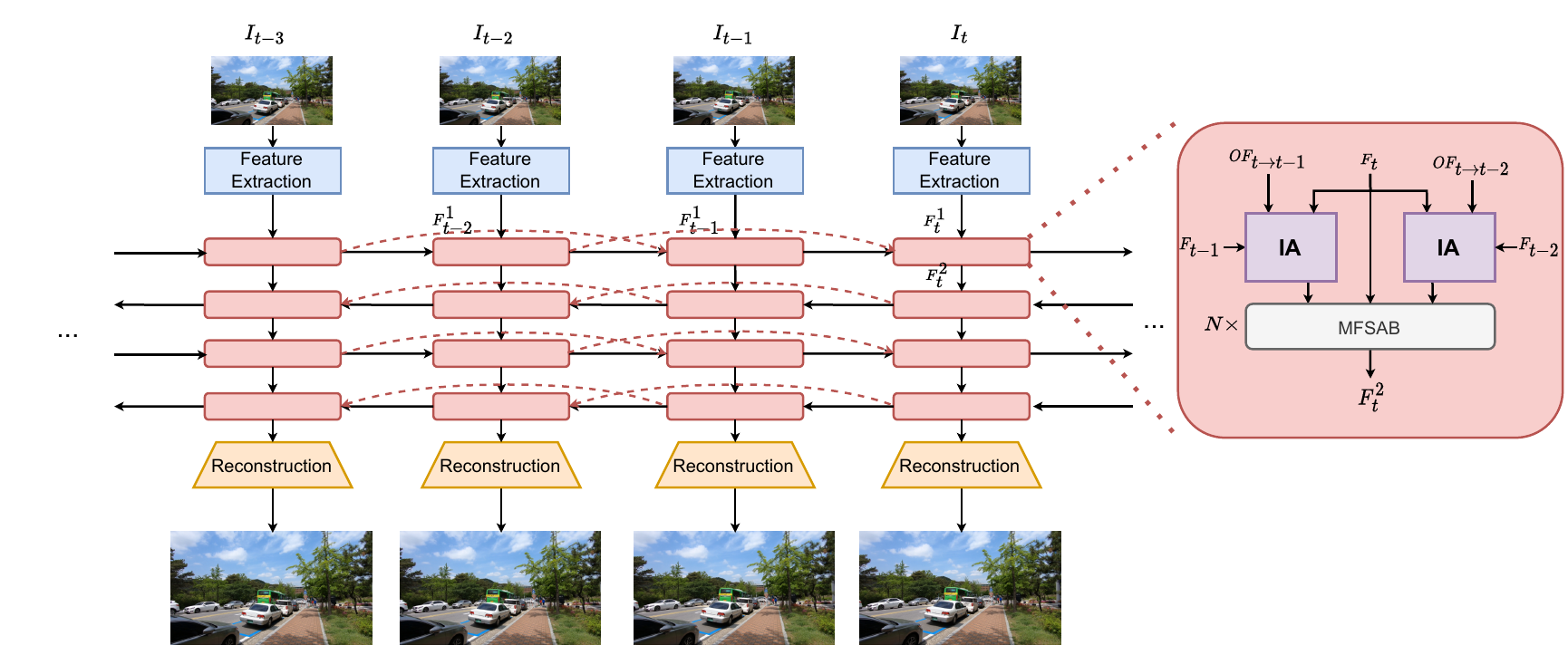}%

\caption{Network Structure for IA-RT.}\label{fig:sup_network_structures_iart}

\end{figure*}

The detailed network structure for IA-CNN is provided in \cref{fig:sup_network_structures_iacnn}, while that for IA-RT is presented in \cref{fig:sup_network_structures_iart}.

For IA-CNN, we adapted first-order bidirectional propagation following BasicVSR \cite{shi2022rethinking}. The number of channels was set to 64, and each propagation branch comprised 30 residual blocks. The IA module had 64 channels, and the attention module for implicit alignment included 8 heads.

For IA-RT, two consecutive second-order bidirectional propagation blocks were employed, following the PSRT-recurrent model \cite{shi2022rethinking}. The number of channels for embedding was 120, and each propagation branch contained 18 Multi-Frame Self-Attention Blocks (MFSABs) \cite{shi2022rethinking} with shortcuts every 6 blocks. The IA module had 120 channels, and the attention module included 6 heads.

\section{Experimental Details}

\subsection{Alignment Study}
For the synthetic data, we partitioned the training videos from the clean data track of Sintel~\cite{butler2012naturalistic_sintel} into 20 training and 3 testing videos, and we present the results based on the testing set comprising \textit{ambush$\text{5}$}, \textit{market$\text{6}$}, and \textit{mountain$_\text{1}$}.

All alignment methods evaluated share a consistent network structure and training parameters. We employed the Adam optimizer for training, running for 100,000 iterations at a learning rate of 2e-4, with a batch size of 8.

For alignment studies on the Sintel dataset, we utilized first-order backward propagation due to the availability of only backward optical flow ground truth. Each propagation branch contains 9 MFSABs with shortcuts every three blocks. The number of channels for embedding is set to 60. The total training iterations are 100,000, and the learning rate starts at 2e-4, with a cosine learning rate decay to 1e-7 towards the end of training. The batch size remains 8 throughout.

\subsection{Comparison with State-of-the-Art:} 
All experiments were conducted using bicubic 4X down-sampling. The training dataset includes the REDS~\cite{nah2019ntire} and Vimeo-90K~\cite{xue2019video_tof} datasets, while the testing dataset comprises REDS4~\cite{nah2019ntire}, Vid4~\cite{liu2013bayesian}, and Vimeo-90K-T~\cite{xue2019video_tof}.

For IA-CNN on the REDS dataset, we trained for 300k iterations using 15 input frames, with a learning rate of 2e-4 and a cosine learning rate decay to 1e-7. The batch size was set to 8. Subsequently, we fine-tuned the model on the Vimeo-90K dataset for 300k iterations using the pre-trained weights from the REDS training.

For IA-RT on the REDS dataset, we trained for 300k iterations using 16 input frames, with a learning rate of 2e-4 and a cosine learning rate decay to 1e-7. The batch size remained 8.

When training IA-RT on the Vimeo-90K dataset, we conducted 300k iterations with 7 input frames, using a learning rate of 2e-4 and a cosine learning rate decay to 1e-7. We initialized the weights using the model trained on the REDS dataset. The batch size remained 8.

Test results for the REDS model are reported on the REDS4 dataset, while test results for the Vimeo-90K model are reported on Vimeo-90K-T and Vid4.

\noindent\textbf{Evaluation metrics:}
We calculate PSNR and SSIM on the RGB channel for REDS4 and Y channel for Vimeo-90K-T and Vid4.

\noindent\textbf{Ablation Studies:} For ablation studies on the REDS dataset, a model with a single second-order bi-directional propagation block is employed. Each propagation branch consists of 9 MFSABs with shortcuts every 3 blocks. The total training iterations are set to 200,000, with a learning rate initialized at 2e-4 and subjected to a cosine learning rate decay, reaching 1e-7 at the end of training. The batch size used for these experiments is 4.

\subsection{FLOPS and Runtime Comparison}
We conducted an analysis of the FLOPs, as presented in \cref{tab:flops}. The overall complexity of a single IA module is 14.93 GFLOPs with an input low-resolution (LR) frame size of $180\times320$. In comparison with PSRT-recurrent, eight propagations with IA are performed per frame on average. Thus, the FLOPs for Implicit Alignment with Recurrent Transformers (IART) is calculated as follows:
\[ \text{FLOPs}_{\text{IART}} = \text{FLOPs}_{\text{PSRT}} + 14.93\ \text{G}\!\times\!8 = 1.62\ \text{T} \]

For a fair runtime comparison, we estimated the inference time for PSRT-recurrent and IART on the same hardware, namely the RTX-A5000.
\begin{table}[t]
\footnotesize
    \vspace{1mm}
    \centering
    \begin{tabular}{l|c|c|c}
    \toprule
        Method & Param. (M) & FLOPs (T) & Runtime (ms) \\
    \midrule
        DUF & 5.8 & 2.34 & - \\
        RBPN & 12.2 & 8.51 & - \\
        EDVR \cite{wang2019edvr} & 20.6 & 2.95 & - \\
        VSRT \cite{cao2021video} & 32.6 & 1.60 & -\\
        VRT \cite{liang2022vrt_vrt} & 35.6 & 1.30 & -\\
        PSRT-recurrent \cite{shi2022rethinking} & 13.4 & 1.50 & 2020$^\dagger$ \\
        IART (ours) & 13.4 & 1.62 & 2105\\
    \bottomrule
    \end{tabular}
    
    \caption{The comparison of the parameters, FLOPs and the runtime for VSR models. {$^\dagger$The runtime is re-estimated on RTX-A5000 GPU for fair comparison.}}    \label{tab:flops}
    \vspace{-1.5em}

\end{table}

\vspace{-0.5em}
\subsection{Training Time}
The training time on an RTX-A5000 for IART is 6.7 ms/iter for 16 input training frames and 2.6 ms/iter for 6 input training frames. In comparison, the training time for PSRT-recurrent on RTX-A5000 is 6.4 ms/iter for 16 input training frames and 2.5 ms/iter for 6 input training frames.
\vspace{-0.5em}

\section{More Visual Comparison}
\begin{figure*}
\centering
\includegraphics[width=0.9\textwidth]{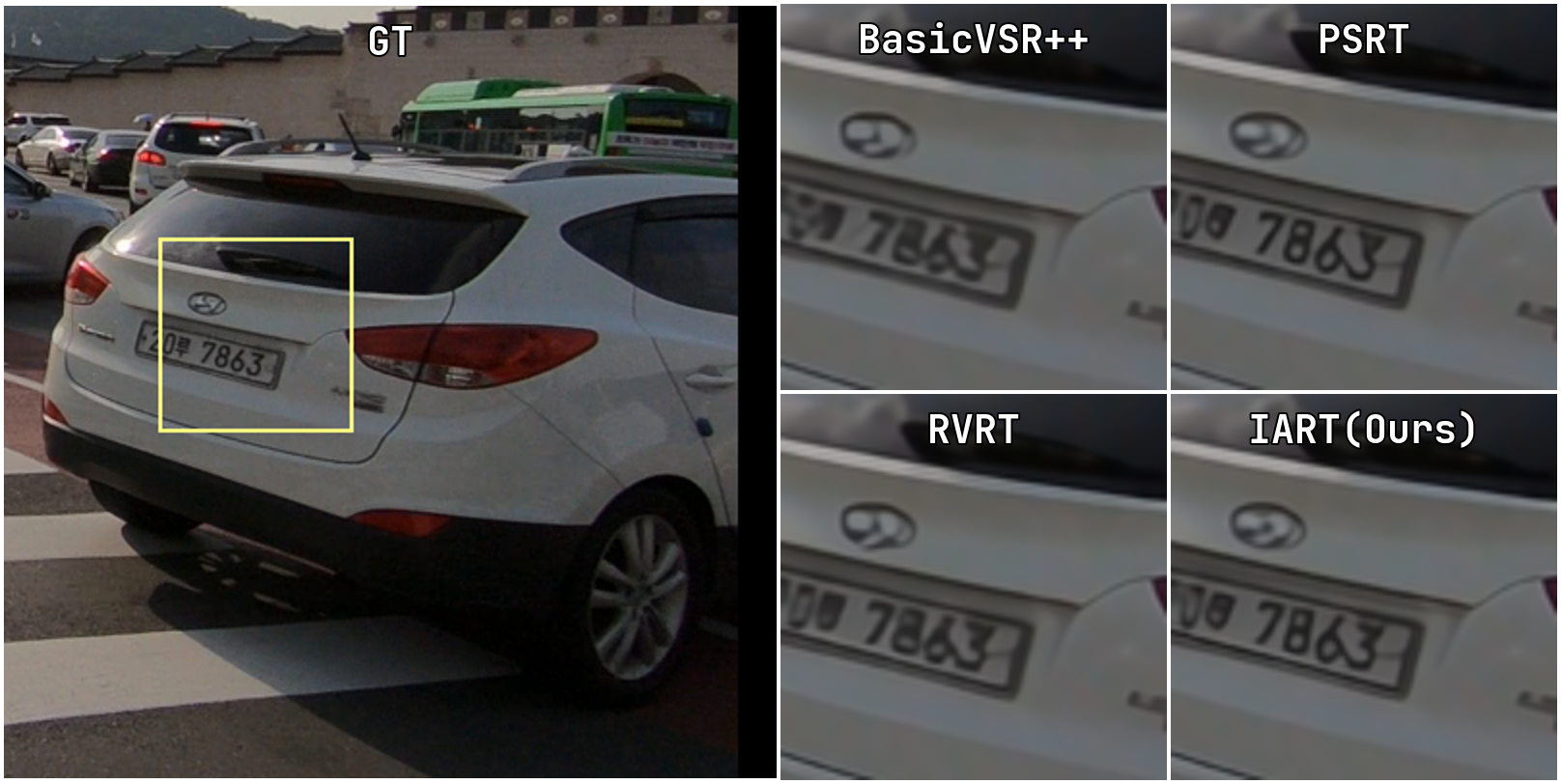}\\
\includegraphics[width=0.9\textwidth]{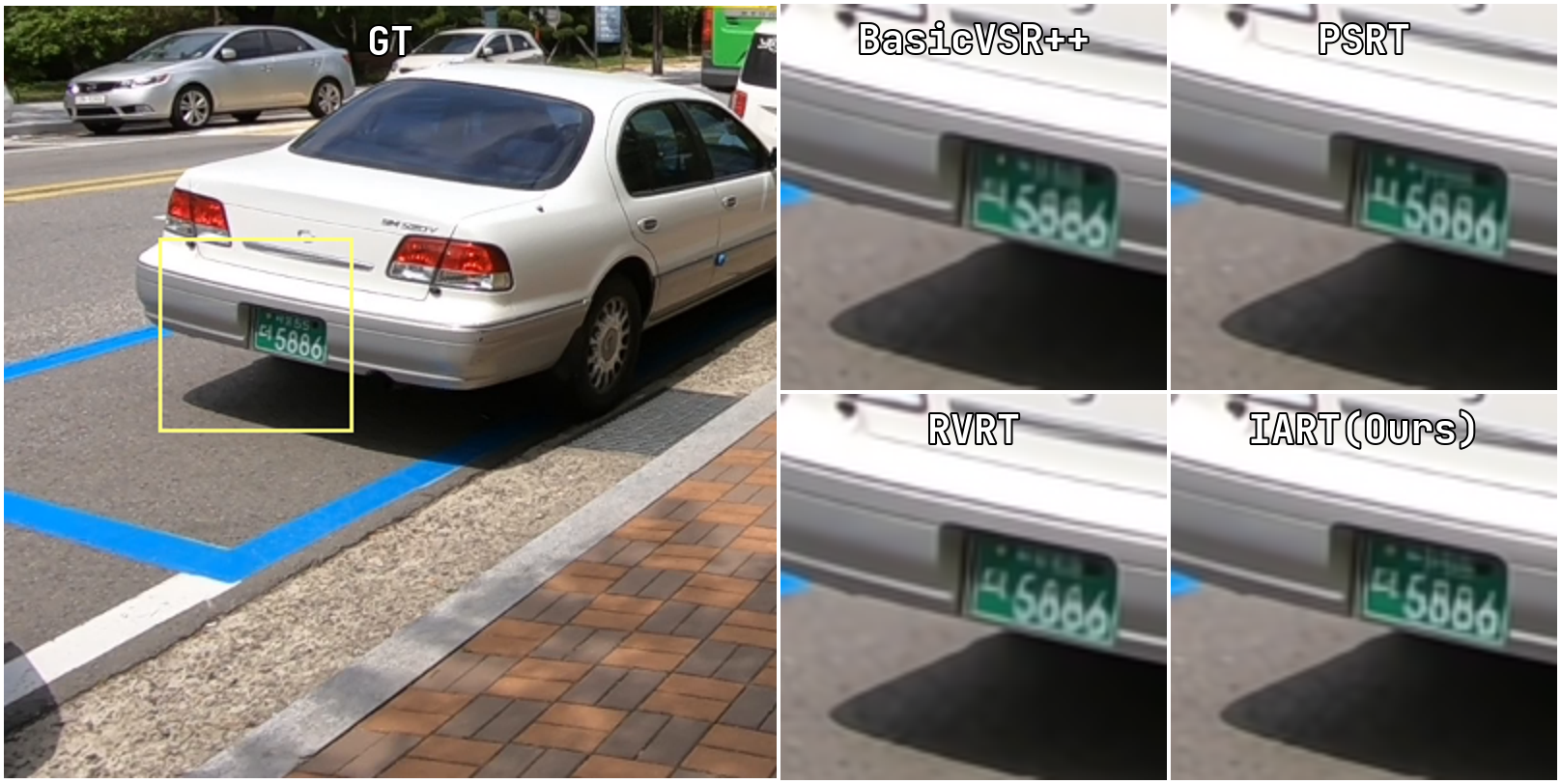}\\

\caption{Visual Comparison on REDS4.}\label{fig:sup_qual1}

\end{figure*}

\begin{figure*}
\centering
\includegraphics[width=0.9\textwidth]{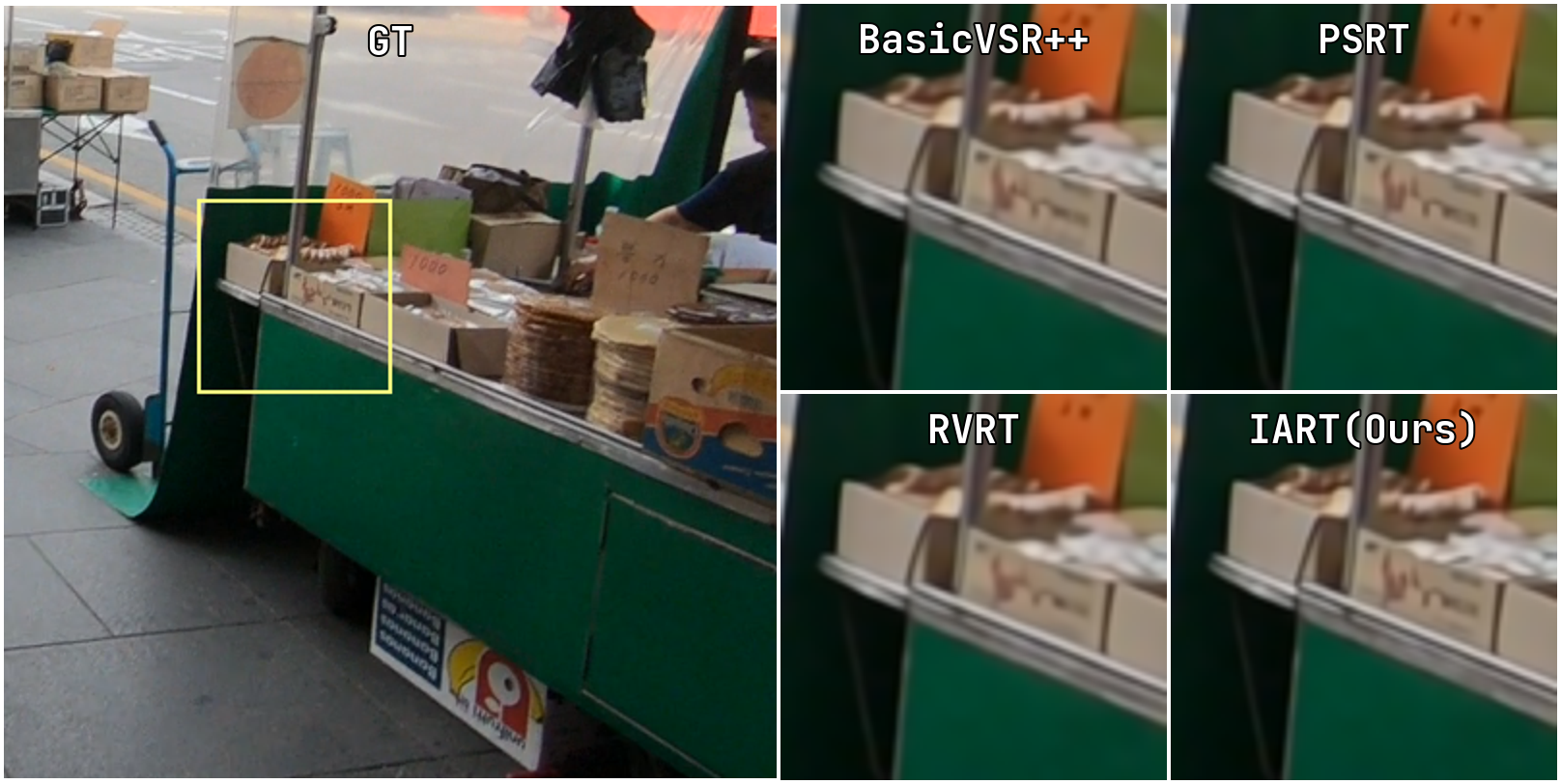}\\
\includegraphics[width=0.9\textwidth]{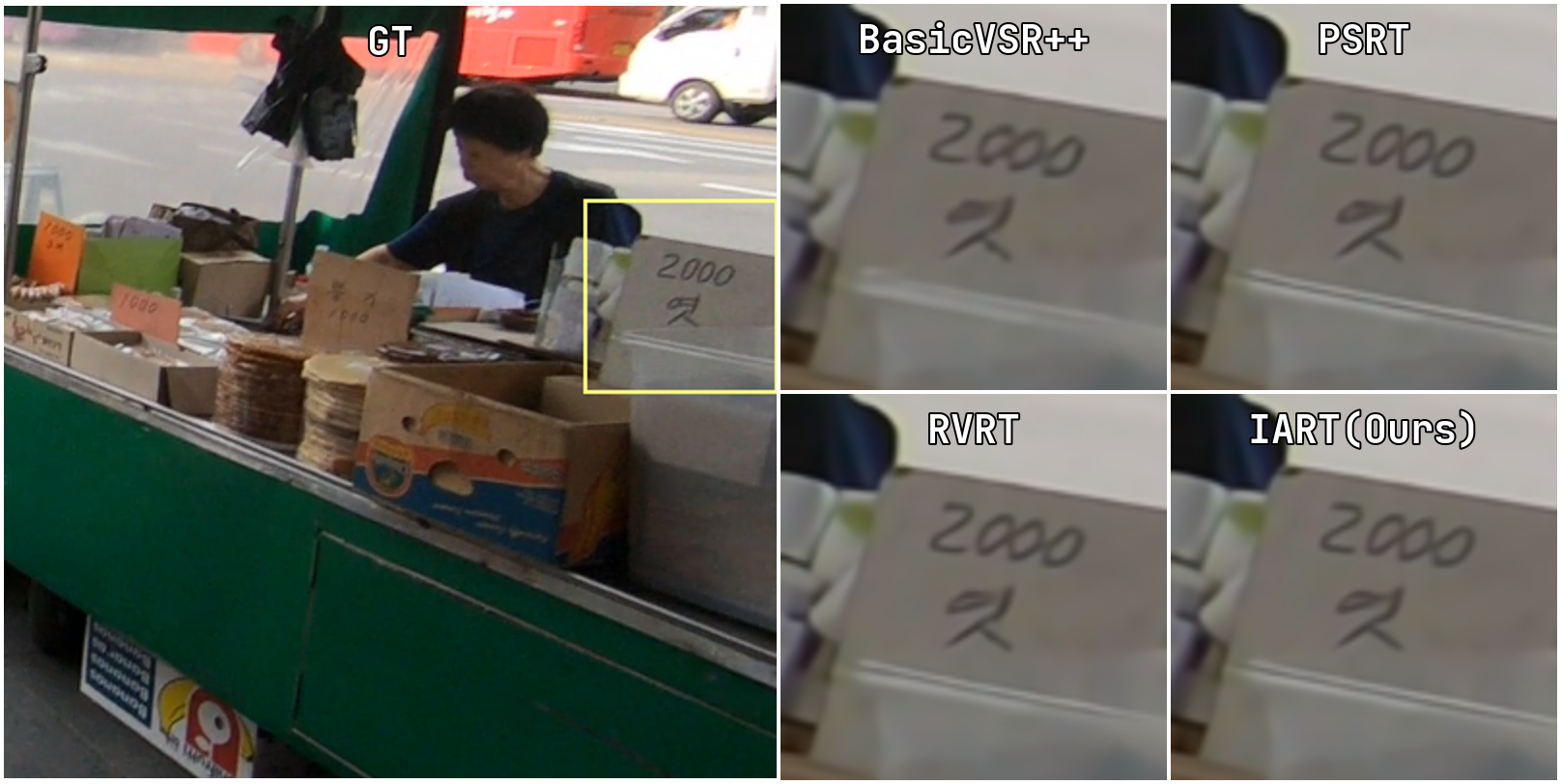}\\

\caption{Visual Comparison on REDS4.}\label{fig:sup_qual2}

\end{figure*}

We offer additional visual comparisons in \cref{fig:sup_qual1} and \cref{fig:sup_qual2}.

{\small

\bibliographystyle{ieeenat_fullname}
\bibliography{main}

}